%% file: main.tex
\documentclass[letterpaper,twocolumn,10pt]{article}
\usepackage{usenix2019_v3}

\usepackage{tikz}
\usepackage{amsmath}

\usepackage{filecontents}

\usepackage{soul, color} 
\usepackage[x11names]{xcolor}
\usepackage{graphicx} 
\usepackage{bbding} 
\usepackage{multirow} 
\usepackage{booktabs} 
\usepackage{makecell} 
\usepackage{colortbl} 
\usepackage{url} 
\usepackage{latexsym} 

\begin{document}

\input{0-title}

\input{1-introduction}
\input{2-background}
\input{3-overview}
\input{4-attention}

\input{5-matmul}

\input{6-evaluation}

\input{7-relatedwork}

\input{8-conclusion}

\bibliographystyle{plain}
\bibliography{reference.bib}

\end{document}

%% file: 0-title.tex
\title{\Large \bf Tackling the Dynamicity in a Production LLM Serving System with SOTA Optimizations via Hybrid Prefill/Decode/Verify Scheduling on Efficient Meta-kernels}

\author{
{\rm Mingcong Song\thanks{These authors contributed equally to this work.}}\\
Huawei
\and
{\rm Xinru Tang\textcolor{Green3}{\footnotemark[1]} }\\
Tsinghua University
\and
{\rm Fengfan Hou}\\
Huawei
\and
{\rm Jing Li}\\
Huawei
\and
{\rm Wei Wei}\\
Huawei
\and
{\rm Yipeng Ma}\\
Huawei
\and
{\rm Runqiu Xiao}\\
Huawei
\and
{\rm Hongjie Si}\\
Huawei
\and
{\rm Dingcheng Jiang}\\
Tsinghua University
\and
{\rm Shouyi Yin}\\
Tsinghua University/Shanghai AI Lab
\and
{\rm Yang Hu}\\
Tsinghua University
\and
{\rm Guoping Long}\\
Huawei
} 

\maketitle

\begin{abstract}
 Meeting growing demands for low latency and cost efficiency in production-grade large language model (LLM) serving systems requires integrating advanced optimization techniques. However, dynamic and unpredictable input-output lengths of LLM, compounded by these optimizations,  exacerbate the issues of workload variability, making it difficult to maintain high efficiency on AI accelerators, especially DSAs with tile-based programming models. To address this challenge, we introduce XY-Serve, a versatile, Ascend native, end-to-end production LLM-serving system. The core idea is an abstraction mechanism that smooths out the workload variability by decomposing computations into unified, hardware-friendly, fine-grained meta primitives. For attention, we propose a meta-kernel that computes the basic pattern of matmul-softmax-matmul with architectural-aware tile sizes. For GEMM, we introduce a virtual padding scheme that adapts to dynamic shape changes while using highly efficient GEMM primitives with assorted fixed tile sizes. XY-Serve sits harmoniously with vLLM. Experimental results show up to $89\%$ end-to-end throughput improvement compared with current publicly available baselines on Ascend NPUs. Additionally, our approach outperforms existing GEMM (average $14.6\%$ faster) and attention (average $21.5\%$ faster) kernels relative to existing libraries. While the work is Ascend native, we believe the approach can be readily applicable to SIMT architectures as well.
    
\end{abstract}

%% file: 1-introduction.tex
\section{Introduction}

\begin{figure}[t]
    \centering
    \includegraphics[width=1.0\linewidth]{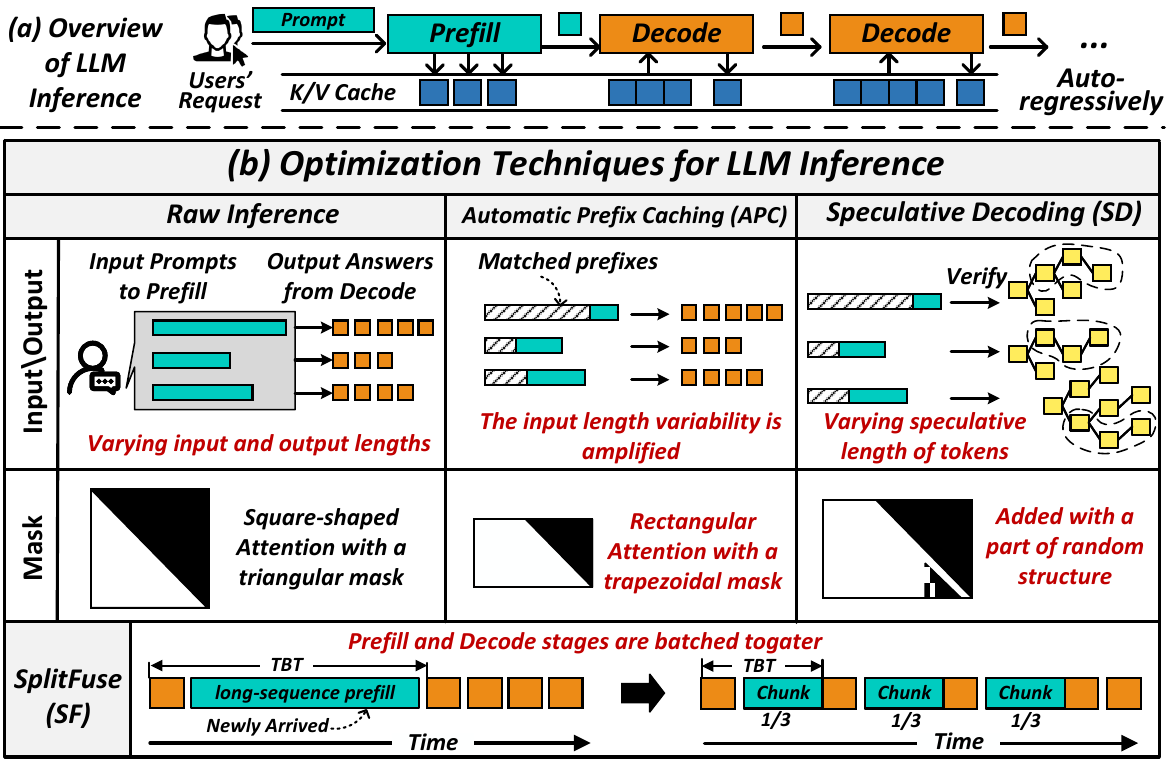}
    \caption{Dynamics of LLM Inference.}
    \label{fig-background}
\end{figure}

Large language models (LLMs)\cite{yang2024qwen2, touvron2023llama} have achieved impressive accuracy and are widely applied in fields like natural language processing\cite{openai_gpt} and computer vision\cite{carion2020end, dosovitskiy2021an}. As shown in Fig. \ref{fig-background}(a), LLM inference typically consists of two stages: prefill and decode. During the prefill stage, LLMs process the user’s input to generate an initial token, while concurrently caching the key/value (K/V) data for future use. In the decode stage, tokens are generated sequentially in an auto-regressive manner. Despite their impressive performance, LLMs come with significant computational costs and latency. As the model size increases and input sequences become longer, the computational demands grow substantially, making online inference increasingly challenging\cite{kaplan2020scaling}. 

To address these challenges, a number of optimization techniques have emerged to reduce inference costs and latency, such as Automatic Prefix Caching (APC)\cite{vLLM_APC, zheng2312sglang}, Speculative Decoding (SD)\cite{cai2401medusa, li2024eagle, miao2024specinfer, zhao2024lookahead, chen2023accelerating, leviathan2023fast}, and SplitFuse\cite{holmes2024deepspeed, agrawal2023sarathi, agrawal2024taming}. APC enables new queries with matching prefixes to reuse cached K/V data and skip computations for shared segments, thereby improving prefill performance. The adoption of draft model in SD provides a chance for target model to generate multiple tokens per step, enhancing key/value cache and model weight reuse, which helps mitigate the memory-bound bottleneck of decode in draft model. SplitFuse splits long-sequence prefill tokens into smaller chunks and schedules them alongside decode tokens, reducing interruptions to the decode stage.

While these optimizations promise to improve inference efficiency, they also introduce new complexities. For LLM serving of a production system, the key challenge is how to integrate all these optimizations efficiently on AI accelerators given that a friendly SIMT programming model is lacked. We illustrate this issue in Fig. \ref{fig-background}(b). Performance already struggles with unpredictable and varying input/output token lengths, these optimizations can further exacerbate the issue. 

For example, APC increases the variability in input prompt lengths, as the number of cached prefixes depends on both query history and real-time memory availability. With SD, the decode stage no longer processes one token at a time. Instead, it handles a dynamically varying speculative length of tokens\cite{liu2024optimizing}. This new stage is referred to as the Verify stage \cite{zhang2023draft, huggingface_verify}. SD also transforms the attention mask from a standard causal mask into a more complex, dynamically generated version. Additionally, SplitFuse combines the prefill and decode stages into a single batch, further complicating the management of multiple stages within one batch.

These dynamicities pose significant challenges for LLM computations, particularly in Linear and Attention modules. First, the uncertainty of input lengths leads to arbitrary matrix shapes in Linear ops, complicating optimization efforts aimed at achieving peak computational efficiency. Second, the adoption of technologies like APC, SD, and SplitFuse introduces greater diversity in attention shapes and mask structures, weakening the effectiveness of existing attention kernels.

Furthermore, in practical systems, the Prefill (P), Decode (D), and Verify (V) stages may operate independently or in combination. Even in disaggregated deployments, the D node may run both D and V stages simultaneously. If disaggregated deployment\cite{hu2024inference, qin2024mooncake, zhong2024distserve, jin2024p} nodes
 support dynamic role switching, the hybrid P/D/V combinations issue may also arise during role transitions. Since these stages present varying computational loads during the attention phase, attempting to enumerate and optimize for every possible stage combination becomes a labor-intensive and impractical task.

To tackle the challenges posed by dynamicities, we present XY-Serve,  a versatile, Ascend native, and end-to-end production LLM-serving system. The main idea is to introduce an abstraction to bridge the gap between varying high level workloads and fixed hardware-friendly low level meta-primitives. Specifically, XY-Serve features a token-wise scheduling mechanism that batches tokens in chunks. Tokens could be from either prefill, verify, and decode stages. Then the token chunks will be processed by three core components: workload decomposition, computation task reordering, and meta kernels.

Workload decomposition is a mechanism to decompose and map dynamic workloads onto hardware-friendly meta-primitives. For attention computations, it unifies the P/D/V stages through dynamic tiling, generating hardware-friendly tile-based computational tasks. Each task computes a basic matmul-softmax-matmul pattern. For GEMM, it decomposes dynamic-shaped matmul ops into a small set of basic matmul primitives with fixed tile sizes. This transformation seems deceptively straightforward but is surprisingly hard to implement without any actual padding overhead. We introduce a novel virtual padding mechanism to address this issue without introducing any extra overhead on AI accelerators with tile-based programming models.

After decomposition, computation task reordering reorganizes decomposed tasks and schedules them for hardware processing cores. The goal of reordering is to smooth out varying task sizes at fine-grained level, balancing workload on different cores for high execution efficiency. For attention, P/D/V tasks have drastically different granularity; thus, reordering of decomposed tasks is essential to achieve load balance. The case for GEMM is different; task reordering is an effective approach to improve L2 cache locality and mitigate bank conflicts. 

With workload decomposition and task reordering, it is possible to construct efficient kernels for attention and GEMM. For attention, we propose meta kernels to efficiently parallelize computations of matmul-softmax-matmul patterns with different tile sizes, without needing to differentiate which P/D/V stage they originate from. With this decoupling approach, it is feasible to integrate a range of optimizations seamlessly and efficiently, including APC, PageAttention\cite{kwon2023efficient}, SplitFuse, SD, and FlashAttention\cite{dao2022flashattention}. These optimizations work synergistically within our meta kernels, achieving superior performance. 

Moreover, our attention meta kernels come with aggressive on-chip Cube Core-Vector Core orchestration pipelines, together with novel schemes to handle various kinds of attention masks dynamically. These low level techniques minimize off-chip memory accesses and maximize on-chip workload balance and execution efficiency. For GEMM ops, we first design and implement a set of highly efficient meta kernels with fixed tile sizes. To handle arbitrary input shapes dynamically, we employ virtual padding at the on-chip memory level, coupled with selective HBM reads and writes, and do not introduce any actual padding overhead. In other words, our approach allows matrix computations to seamlessly handle dynamic shapes while preserving the performance benefits of fixed-shape optimizations.

Experimental evaluation shows that our attention kernels achieve higher efficiency in production workloads, outperforming existing implementations(torch-npu PFA\cite{kernel_PFA} and IFA\cite{kernel_IFA}) by on average of $21.5\%$. Our GEMM kernels not only support arbitrary matrix shapes but also improve performance by on average of $14.6\%$ over existing baselines. In end-to-end evaluation, XY-Serve achieves improvement up to $89\%$ over Ascend-vLLM\cite{vLLM-for-NPU} on publicly available datasets\cite{nightly-benchmark}. While currently implemented on Ascend NPUs\cite{liao2021ascend, liao2019davinci}, these techniques can be applied to other AI platforms as well. Finally, we conduct a comparative evaluation against GPU-based inference systems. In terms of end-to-end MFU and MBU, XY-Serve performs on par with the best GPU implementations.


%% file: 2-background.tex
\section{Motivation}





\begin{figure}[t]
    \centering
    \includegraphics[width=1.0\linewidth]{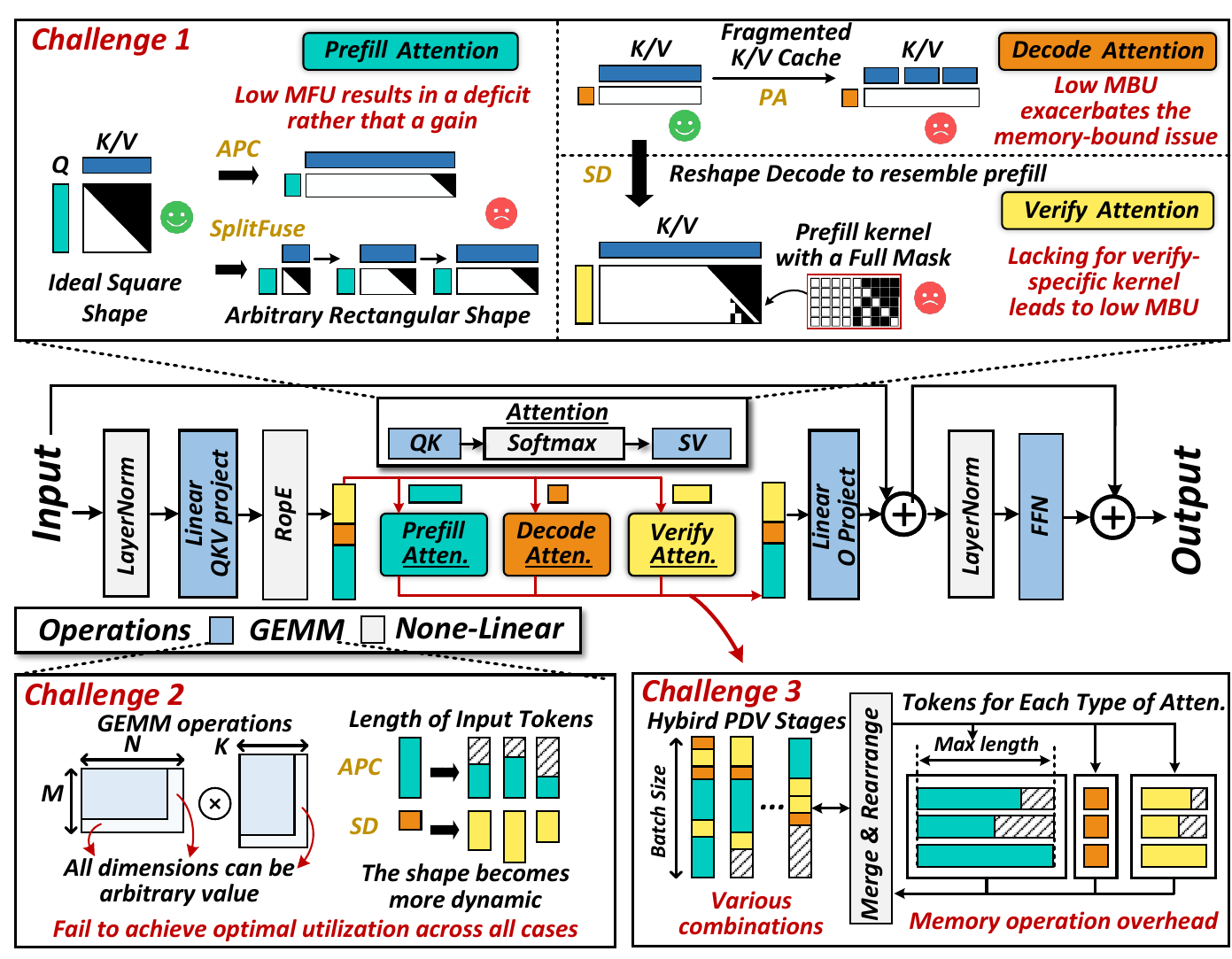}
    \caption{Challenges Posed by Dynamic Workloads.}
    \label{fig-challenges}
\end{figure}

In this section, we explore the challenges posed by dynamic workloads in Linear and Attention modules and analyze the complexities of handling P/D/V hybrid stages, shown in Fig. \ref{fig-challenges}. Finally, we discuss the additional challenges faced by AI accelerators with tile-based programming models, such as the Huawei Ascend NPU\cite{liao2021ascend, liao2019davinci}, in supporting dynamic workloads.

\subsection{\textbf{Diverse Attention}}
\begin{figure}[t]
    \centering
    \includegraphics[width=0.99\linewidth]{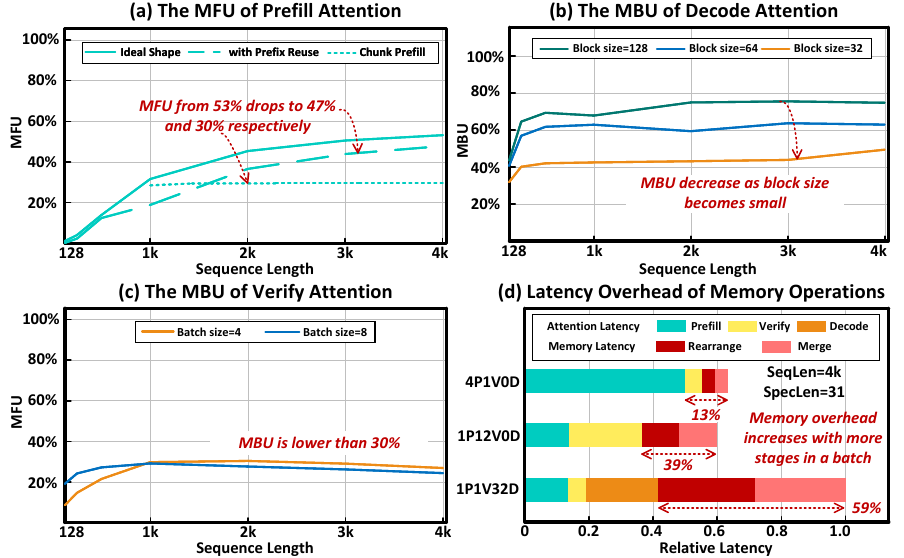}
    \caption{The MFU and MBU of Attention Kernel.}
    \label{fig-moti-kernel}
\end{figure}

The introduction of new technologies, such as APC, SD, and SplitFuse, increases the diversity of Attention shapes and mask structures, leading to MFU and MBU issues of the current NPU attention kernel (torch-npu 2.1 FusedInferAttentionScore \cite{kernel_IAS}), as illustrated in Fig \ref{fig-moti-kernel}.

Firstly, for prefill Attention, without any optimizations, the query length equals the key-value length, resulting in a square-shaped Attention score matrix and a lower triangular mask. This shape is ideal for optimization\cite{lin2024fastattention}, as the sparsity in the mask can be exploited to achieve high performance. The current state-of-the-art (SOTA) torch-npu kernel efficiently handles such square-shaped inputs with triangular masks, achieving an MFU of $53\%$.

However, in real-world scenarios involving prefixes, parts of the key and value are reused from the K/V cache\cite{vLLM_APC, zheng2312sglang}, transforming the Attention score shape from a square to a rectangle with arbitrary dimensions. For these rectangular shapes, the performance of the torch-npu kernel degrades significantly, with MFU dropping to $47\%$ and $30\%$, respectively. While the reuse of the K/V cache theoretically reduces computation, the decline in kernel efficiency offsets this benefit, resulting in no meaningful end-to-end performance improvement.

Secondly, unlike Linear, where tokens from different batches can share weights, decode Attention requires each token to have its own independent K/V cache, making reuse impossible. This leads to inherently memory-bounded computations. PagedAttention (PA)\cite{kwon2023efficient} further fragments the K/V cache access patterns, limiting access to small block-sized portions instead of large contiguous blocks. As illustrated in Fig. \ref{fig-moti-kernel}(b), the MBU decreases as block size becomes small, exacerbating the memory bottleneck of the decode stage.

Lastly, Speculative Decoding (SD)\cite{cai2401medusa, li2024eagle, miao2024specinfer} reshapes decode Attention to resemble prefill Attention but with a shorter query length. Different SD algorithms generate tokens with varying causal structures, resulting in diverse masks during the verify stage. However, due to the absence of a verify-specific kernel for Ascend, the prefill Attention kernel—equipped with full masks—is used as a fallback. This fallback prevents verify-specific optimizations, leading to MBU lower than $30\%$.

\begin{figure}[t]
    \centering
    \includegraphics[width=1.0\linewidth]{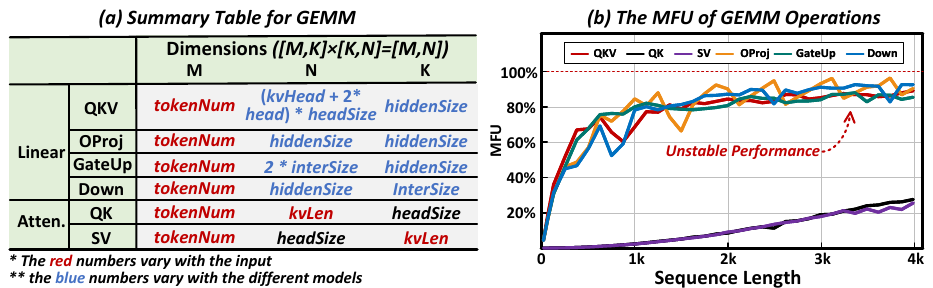}
    \caption{GEMM Operations in LLMs and their MFU.}
    \label{fig_moti_gemm}
\end{figure}

\subsection{\textbf{Dynamic GEMM}}

GEMM is a core operation in LLMs. Beyond the four GEMM operations in the Linear section, the query-key ($QK$) and score-value ($SV$) computations in Attention also rely on GEMM. We summarize all of GEMM operations in Fig. \ref{fig_moti_gemm}(a). For Linear operations, the $M$ dimension is tied to the number of tokens. For Attention $QK$ and $SV$ operations, the $N$ and $K$ dimensions are dependent on the token K/V length. As previously discussed, the number of tokens in the prefill and verify stages is highly dynamic, and this variability is further amplified by the introduction of Automatic Prefix Caching (APC)\cite{vLLM_APC, zheng2312sglang}, making token lengths even more unpredictable.

In addition, for Linear GEMM, the dimensions $N$ and $K$ are influenced by  $hiddenSize$, which varies with the architecture of the LLM. Consequently, the $M$, $N$, and $K$ dimensions in GEMM operations for LLM are all dynamically changing.

Designing a matrix multiplication on AI accelerators to support arbitrary shapes while ensuring high performance is inherently challenging. Most accelerators are typically optimized for specific tile sizes, such as $16\times 16$, making it difficult to fully utilize their capabilities with arbitrary shapes. Irregular shapes complicate parallelism and load balancing, causing inefficiencies in workload distribution across processing units. The requirement for flexible algorithms to adapt to diverse shapes increases algorithmic complexity. Additionally, handling boundary conditions for non-uniform matrix sizes introduces overhead, further affecting performance. As shown in Fig. \ref{fig_moti_gemm}(b), employing a general-purpose GEMM kernel (torch-npu 2.1 linear operator\cite{_torchnnnative_}) to accommodate all possible results in unstable performance and fails to achieve optimal utilization across all GEMM operations.

\subsection{\textbf{Hybird P/D/V Stages}}

In practical systems, P/D/V stages may exist independently or coexist simultaneously, leading to arbitrary combinations of interleaved P/D/V stages within a given scheduling budget. For Linear operations, tokens from different stages can be grouped together and treated as the left matrix in a GEMM operation, sharing the same weight matrix on the right. This approach is straightforward, as tokens from different stages can reuse the same large model weights.

In contrast, handling Attention operations is significantly more complex. Stages are independent, and even within the same stage, batches are also independent. Enumerating and tailoring optimizations for each possible combination of stages and batches is highly labor-intensive and impractical.

A common alternative is a batch-by-batch execution, such as selective batching\cite{yu2022orca}, which processes each batch independently by invoking the corresponding Attention kernel. However, this method introduces additional memory overhead from splitting, rearranging, and merging data, which degrades overall system performance. As shown in Fig. \ref{fig-moti-kernel}(d), the memory overhead may account for more than $50\%$. Furthermore, batch-by-batch processing leads to inefficient utilization of computational resources, further limiting system efficiency.

\subsection{Ascend NPU Micro-architecture}

\begin{figure}[t]
\centering
\includegraphics[width=1.0\linewidth]{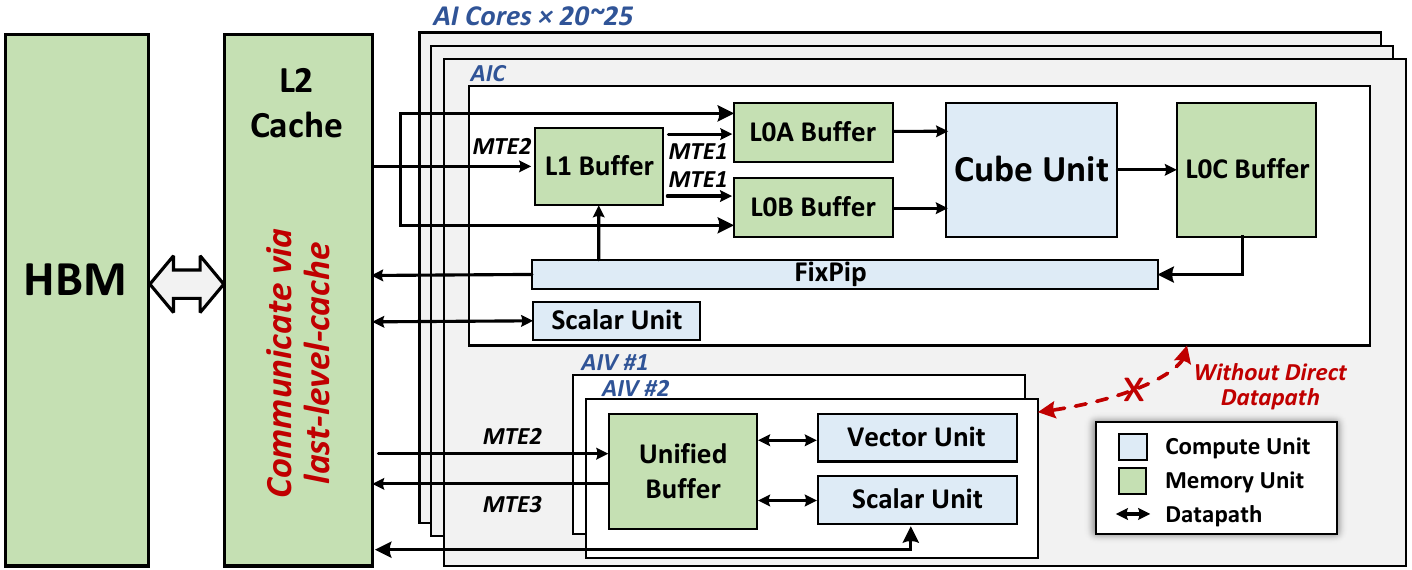}
\caption{The Micro-architecture of Ascend 910B.}
\label{fig-ascend}
\end{figure}

Built on Huawei's DaVinci architecture \cite{liao2019davinci, liao2021ascend}, the Huawei Ascend NPU is a high-performance AI processor. Fig. \ref{fig-ascend} illustrates the micro-architecture of the Ascend, which consists primarily of AIC and AIV components. The AIC, similar to Nvidia's Tensor Core, handles matrix computations, while the AIV is responsible for vector operations. AIC and AIV are separated without a direct datapath, so ensuring their data interactions occur via the L2 cache is crucial when designing mixed kernels. Compared to GPUs, NPUs have larger core granularity, making load balancing between cores even more critical. The Memory Transfer Engine (MTE) handles data movement. Whether AIC, AIV, or MTE, all data processing and transferring occur at the tile level. AI accelerators with tile-based programming models are gaining increasing attention in the community. However, this tile-based processing model encounters significant challenges when managing dynamic workloads that vary at the token granularity. Typical solutions involve complex padding/unpadding operations, which result in wasted computation and memory, leading to substantial overhead.

%% file: 3-overview.tex
\section{Overview of XY-Serve}

To address the aforementioned challenges, we developed XYServe, a versatile end-to-end production LLM-serving system, which is built on four key components: Token-wise Scheduling, Dynamic Task Decomposition and Reordering, Meta-Attention, and SmoothGEMM.

\subsection{Token-wise Scheduling}
\label{sec-overview}

\begin{figure*}[h]
    \centering
    \includegraphics[width=1.0\linewidth]{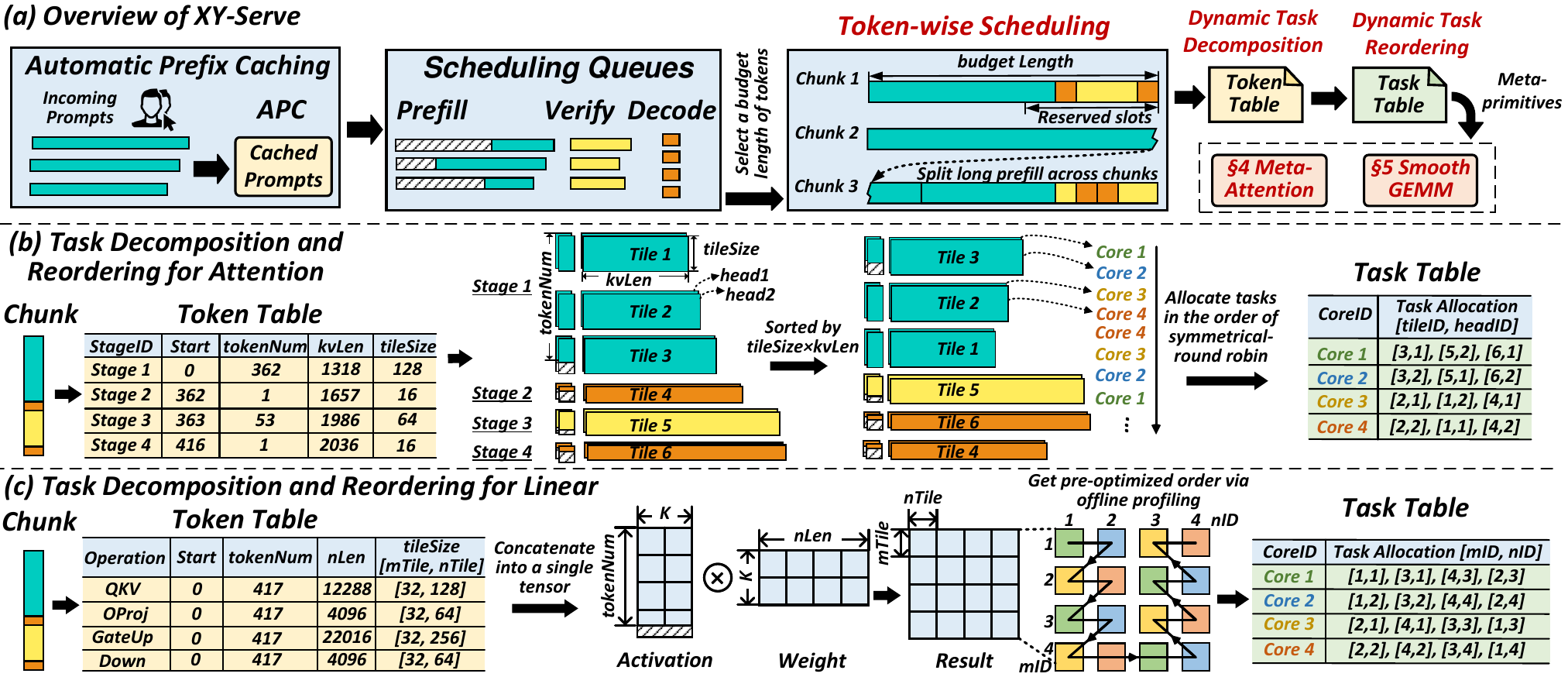}
    \caption{Overview of XY-Serve.}
    \label{fig-overview}
\end{figure*}

When a user's request enters the system, it first passes through the APC module, which matches the incoming prompt against existing prompts in the K/V cache, enabling token-wise reuse. Any unmatched tokens are added to the scheduling queues. Consequently, the prompt length in the scheduling queues is the user's input length minus the length of tokens already cached in the K/V cache. Since both the user's input and the cached token lengths are dynamically variable, the prompt lengths in the scheduling queues become even more dynamic. The scheduling queues also include decode and speculative tokens from previous requests awaiting processing.

Our scheduling system selects a fixed-budget length of tokens from the scheduling queues to form chunks, which may include tokens from prefill, verify, or decode stages. To improve first-token latency, prefill requests are prioritized. If a user's prefill prompt exceeds the budget length, it is split into smaller parts to ensure each scheduled chunk remains within the budget. To minimize interruptions caused by prefill on decode and maintain a stable Time Between Tokens (TBT), certain slots are reserved for decode and speculative tokens. Prefill-only chunks are scheduled only when the queue has no decode or speculative tokens.

As shown in Fig. \ref{fig-overview}(a), the composition of P/D/V stages within a single chunk is inherently unpredictable, with token counts for each stage varying arbitrarily and each stage having a distinct historical K/V length. Additionally, the total token count in a chunk may not always match the budgeted length and can fall below the budget under a low system load. To address the four levels of dynamism, our scheduling operates entirely at the granularity of individual tokens without concern for their origin.

\subsection{Task Decomposition}

While token-wise scheduling improves efficiency by reducing bubbles and optimizing resource utilization, the four levels of dynamism it introduces pose significant challenges for execution, especially on AI accelerators with tile-based programming models.

To address these challenges, we propose a dynamic decomposition mechanism that converts dynamic workloads into hardware-friendly, tile-based computational units. Using the Token-Table, each stage is logically decomposed into tile blocks. At the tile level, computation modules can process these blocks in parallel without distinguishing their P/D/V stage origin. Importantly, this tiling decomposition is purely logical, requiring no changes to the physical data layout.

In the following sections, we present a detailed analysis of how tiling decomposition is applied to Attention and Linear layers.


\subsubsection{Attention Decomposition}
For attention, the Token-Table contains entries for each P/D/V stage, with each entry specifying key attributes, including the \emph{stageID}, the \emph{start position}, the number of \emph{tokenNum}, the historical \emph{kvLen}, and the \emph{tileSize}. As shown in Fig. \ref{fig-overview}(b), stage-1 (P) is decomposed into three tiling blocks, while stage-2 and stage-4 (D) are each divided into one tiling block, and stage-3 (V) is also decomposed into one tiling block. Each tiling block consists of $headNum$ tiling units, resulting in a total of $6 \times headNum$ tiling units at the tiling level.

\subsubsection{Linear Decomposition}
For Linear operations, since tokens from different stages can share the same weights, they are concatenated into a single, large tensor and multiplied by the shared weights. This approach avoids multiple GEMM invocations, enhances weight reuse, and streamlines computation. Consequently, Linear operations do not need to be differentiated between stages, and can be processed uniformly.

In the Token-Table, each Linear operator corresponds to a single entry shown in \ref{fig-overview}(c). The four primary Linear operations are \emph{QKV}, \emph{OProj}, \emph{GateUp}, and \emph{Down}. For these operations, the \emph{start position} is set to $0$, indicating that all tokens are processed from the beginning of the concatenated tensor. The \emph{tokenNum} equals the total number of tokens in the currently scheduled chunk. Tiling is performed on the result matrix of dimensions $tokenNum \times nLen$, where each tiling block corresponds to a submatrix of the result. Each tiling block is assigned to a single AI core, allowing different cores to process distinct blocks in parallel. Each Linear operator has its own specific $tileSize$, determined by the $tokenNum$ and $nLen$ shapes of the operation.

\subsection{Task Reordering}
After decomposing the dynamic workloads from P/D/V mixed stages into fundamental tile units, it is necessary to reorder these tile units and generate a Task-Table to enhance performance. The Task-Table is responsible for scheduling these tile units onto the hardware, with each entry specifying a \emph{coreID} and the list of tiles assigned to that core. Based on this Task-Table, Attention, and Linear can simply retrieve the corresponding tile units according to their \emph{coreID}. This approach not only maximizes hardware efficiency but also simplifies the design of Attention and Linear kernels.

\subsubsection{Attention Reordering}
After performing dynamic tiling on the various stages, the resulting tiles exhibit varying values for $tileSize$ and $kvLen$. This variation can lead to load imbalances during parallel processing. To address this, we calculate the computational load of each tile as its area, defined as $tileSize \times kvLen$.

For efficient scheduling, the tiles are initially sorted based on their computational load, from largest to smallest. Subsequently, the tiles are allocated to the AI cores in a symmetrical round-robin fashion. As depicted in Fig. \ref{fig-overview}(b), assuming there are four AI cores, the tile units are assigned in the sequence core-1, core-2, core-3, core-4, core-4, core-3, core-2, core-1, and so forth, repeating this pattern to ensure a balanced and efficient allocation of computational tasks.

This task scheduling information is stored in the Task-Table and passed to the Attention module. The Task-Table guides the Attention module to perform parallel processing efficiently, leveraging both the head and tile dimensions. This mechanism ensures balanced computation across AI cores, maximizing hardware utilization.

\subsubsection{Linear Reordering}
Given that only a limited set of fixed shapes is supported, we perform offline optimization to determine the most efficient task allocation strategies for linear ops. This involves profiling and customizing task allocation for each shape to maximize performance. The optimized strategies are stored for use during execution. During runtime, XY-Serve leverages the Token-Table to identify the current shape and uses this information to retrieve the corresponding pre-optimized Task-Table. The Task-Table is then passed to the Linear module, guiding it to execute tasks in an optimized manner.

%% file: 4-attention.tex
\section{Meta-Attention}
\label{sec_attention}

In this section, we first explain how our attention module supports advanced features such as APC, Chunked Prefill, and SD. Then, we describe how we optimize attention performance to push it to the hardware limits.

\subsection{Meta-Attention Design}
\subsubsection{Handling Token-wise Processing}

\begin{figure}[t]
    \centering
    \includegraphics[width=0.95\linewidth]{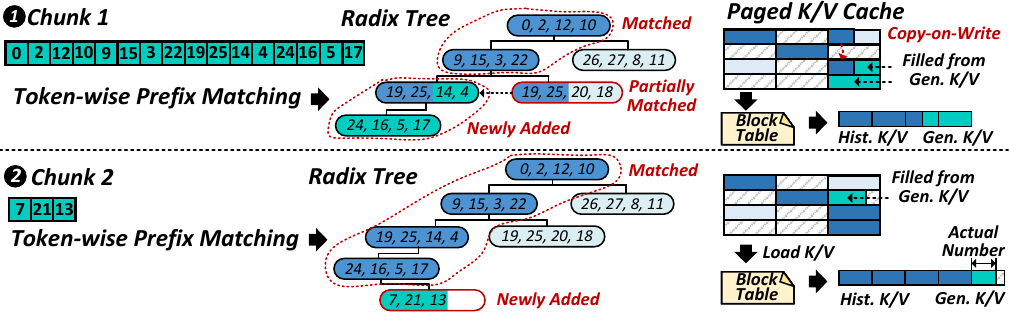}
    \caption{Token-wise K/V Cache Reuse.}
    \label{fig:APC}
\end{figure}

The core requirement for supporting both prefix reuse and Chunked Prefill is that the attention module must be capable of handling arbitrary K/V cache lengths and performing token-wise K/V cache reuse. To achieve this, we use a radix tree to efficiently manage the K/V cache. As shown in Fig. \ref{fig:APC}, each node in the radix tree represents a K/V cache block. The radix tree allows for quick matching of historical K/V cache blocks. When a mismatch occurs at a particular node (i.e., its corresponding K/V cache block contains only a partial match), we use a copy-on-write mechanism to create a new block, refresh the new data into this block, and then add the block back into the radix tree. If additional new blocks are generated after this mismatch, these blocks are directly inserted as child nodes of the mismatched block in the radix tree.

This mechanism effectively manages both historical and newly generated K/V data, seamlessly merging them using copy-on-write to ensure the continuity of the K/V cache. During the prefill attention process, the corresponding K/V blocks—both historical and newly generated—are read based on the block table. We also track the actual number of tokens in the last block, ensuring accurate token-wise processing.

Additionally, our system can automatically cache historical K/V data as prefixes, relieving the user from manually specifying them. The system allows users to set an upper limit on the amount of historical K/V data to be cached. Once this limit is reached, or when space is needed for new K/V data, the system will automatically evict the least recently used K/V blocks from the leaf nodes of the radix tree.


\subsubsection{Minimizing Mask for Speculative Decoding}

Speculative execution can be classified into two types: sequence-based speculation\cite{cai2401medusa, li2024eagle, miao2024specinfer, zhao2024lookahead} and tree-based speculation\cite{chen2023accelerating, leviathan2023fast}. Sequence-based speculation generates multiple tokens within a single sequence; however, its acceptance rate is generally low. In contrast, tree-based speculative algorithms generate predictions for multiple sequences simultaneously, organizing them in a tree structure. This approach can further improve the acceptance rate of speculation.

Both types need support for arbitrary speculation lengths. Furthermore, tree-based speculation utilizes a more complex and dynamic mask, which is not well-suited for vector-based hardware. By analyzing the structure of this mask, we can identify regularities that enable efficient processing.

\begin{figure}[t]
    \centering
    \includegraphics[width=0.95\linewidth]{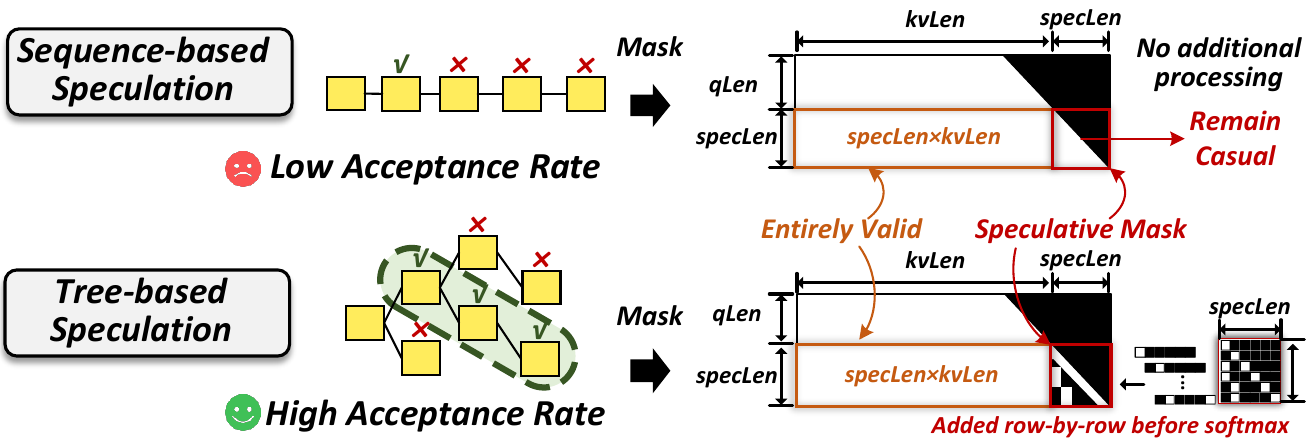}
    \caption{Speculative Decoding Algorithms.}
    \label{fig:sd_mask}
\end{figure}

As illustrated in Fig \ref{fig:sd_mask}, the speculative decoding extends the causal mask of standard prefill ($qLen \times kvLen$) by introducing a $specLen \times (kvLen + specLen)$ region. Within this region, the $specLen \times kvLen$ is entirely valid, while only the $specLen \times specLen$ section requires special handling, referred to as the `Speculative Mask'.

For sequence-based speculation, the speculative mask remains causal, and no additional processing is required, allowing the direct application of our mask-free approach. In tree-based speculation, we generate only the $specLen \times specLen$ part of the mask externally, which is then passed to the kernel and applied to the corresponding attention score matrix.

Our design processes the speculative mask row-by-row, enabling precise control over the start position and length of the mask for each row. This method efficiently supports arbitrary speculation lengths. Once the mask is adjusted, the subsequent computation follows the standard prefill process. Using the speculative mask as a mediator, we can efficiently and seamlessly support a wide range of speculative algorithms.

\subsection{Meta-Attention Optimizations}





\subsubsection{Tile-Based Cube-Vector Orchestration}

\begin{figure}[t]
    \centering
    \includegraphics[width=0.99\linewidth]{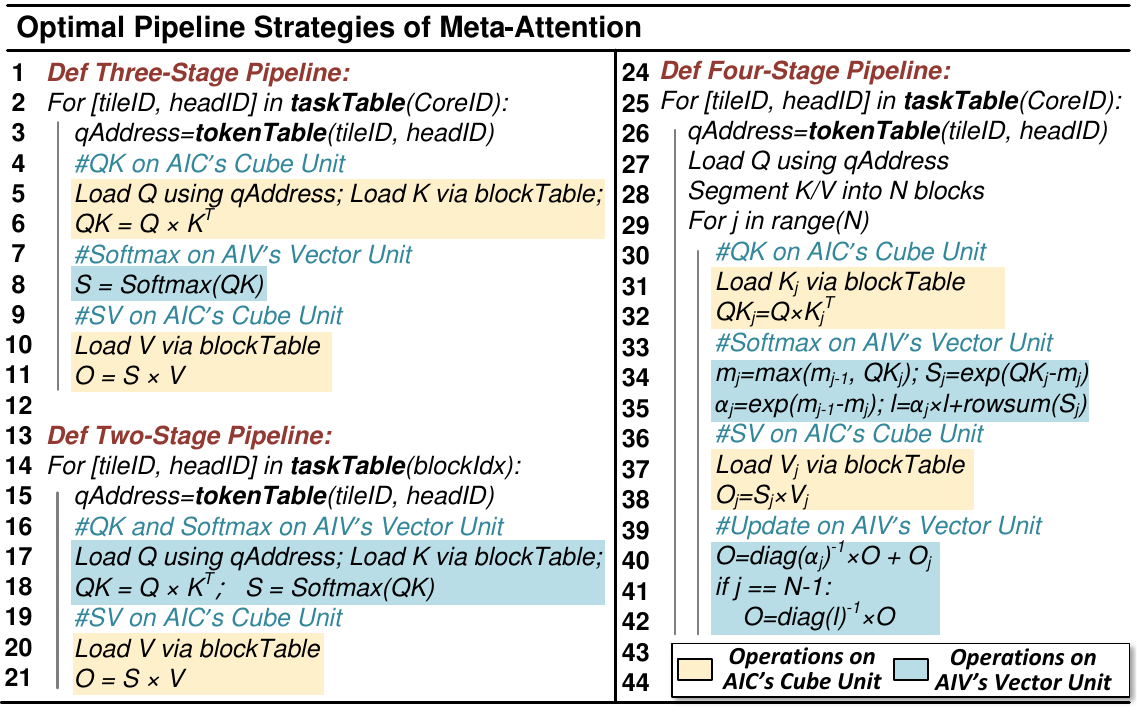}
    \caption{The Algorithm of Cube-Vector Orchestration.}
    \label{fig:FA}
\end{figure}

\begin{figure}[t]
    \centering
    \includegraphics[width=0.95\linewidth]{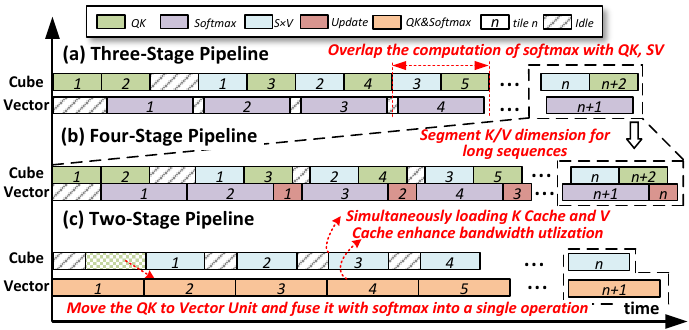}
    \caption{Pipeline of Meta-Attention.}
    \label{fig:pipeline}
\end{figure}

To achieve parallel execution of cube and vector units, we propose a pipeline, shown in Fig. \ref{fig:FA}. It ensures intermediate data transfers occur exclusively via the L2 cache, avoiding costly HBM accesses. The cube unit is responsible for the $QK$ and $SV$ computations, while the vector unit performs the $Softmax$. If processed sequentially ($QK$ → $Softmax$ → $SV$), only one unit would be active at a time, leading to inefficiencies. To address this, we adopted a pipelined approach as displayed in Fig. \ref{fig:pipeline}(a): after completing the $QK$ computation for the first tiling data, its $Softmax$ computation is initiated while simultaneously starting the $QK$ computation for the second tiling data. This overlapping ensures that cube and vector units work concurrently, maximizing hardware utilization.

The intermediate data size for each tile is $tileSize \times kvLen$. When processing $coreNum$ tiles simultaneously, the total intermediate buffer size required is $pipeDepth \times tileSize \times kvLen \times coreNum$. By ensuring this buffer fits within the L2 cache, we can avoid accessing HBM. If the sequence length is short and the intermediate results fit into the L2 cache, there is no need to split along the K/V dimension. In such cases, the three-stage pipeline can be employed, avoiding additional computation and updates required for sequence splitting.

For extremely long sequences, however, splitting along the K/V dimension becomes necessary due to L2 cache limitations. This introduces an additional computation stage, where the $Softmax$ operation is divided into two steps: $Softmax$ and \emph{Update}. The pipeline expands to four stages: $QK \to Softmax \to SV \to$ \emph{Update}. As shown in Fig. \ref{fig:pipeline}(b), we initially prefetch the $QK$ and $Softmax$ computations for one tile of data. Subsequently, cube scheduling alternates between $QK$ and $SV$ tasks, while vector scheduling alternates between $Softmax$ and \emph{Update} tasks. This design enables parallel operation of the cube and vector units, maximizing hardware resource utilization and ensuring optimal performance for both short and long input sequences.

For fully decode-based tasks, we can adopt a new pipeline design to optimize performance further and address the memory-bound issue of decode. Since the Query in decode stage consists of only a single token, the $Query$ matrix is reduced to a vector. Consequently, the $QK$ and $SV$ computations transition from general matrix operations to matrix-vector operations. Executing these operations on cube unit would lead to inefficient utilization of resources. To address this, we move the $QK$ operation to the vector unit and fuse the $QK$ and $Softmax$ operations into a single operator. This fused operator ensures that the execution time for the vector unit aligns closely with the time required for the cube unit to perform the $SV$ operation, effectively balancing the workload.

As illustrated in Fig. \ref{fig:pipeline}(c), the pipeline diagram shows that while the vector unit performs the $QK$ and $Softmax$ operations for tile $n$, the cube unit concurrently executes the $SV$ operation for \emph{tile n-1}. Furthermore, the cube and vector units can simultaneously access the K/V cache data stored in HBM, improving memory bandwidth utilization and further boosting overall performance.

\begin{figure}[t]
\centering
\includegraphics[width=0.90\linewidth]{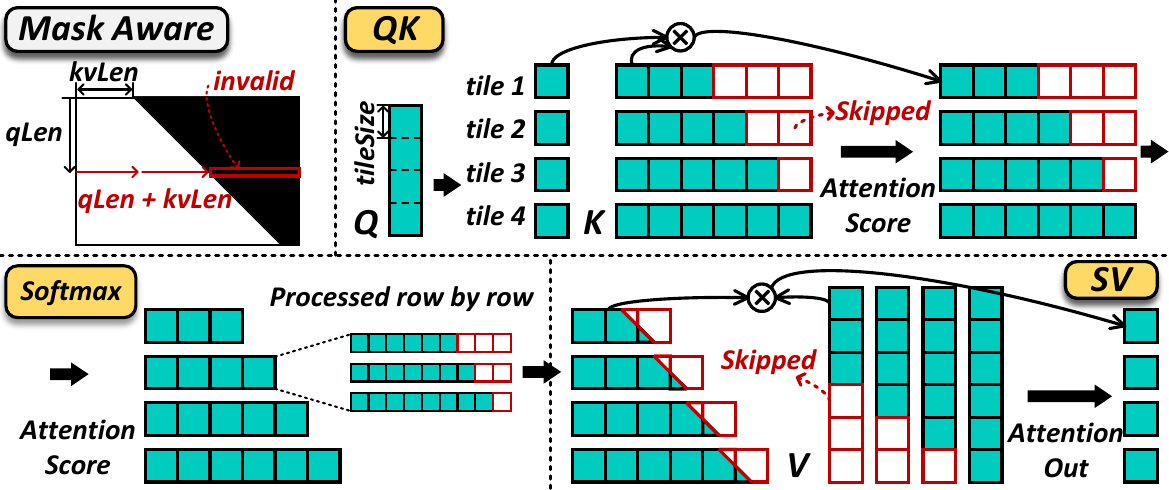}
\caption{Mask-aware Computation.}
\label{fig-casual}
\end{figure}

\subsubsection{\textbf{Exploiting Mask Sparsity}}

To fully utilize the sparsity in the attention mask and skip redundant computations, we adopt a mask-aware strategy that significantly enhances efficiency. As displayed in Fig. \ref{fig-casual}, in the attention computation, once the query dimension coordinate index $qLen$ and the $kvLen$ of K/V are known, any data corresponding to positions after $qLen+kvLen$ in each row of the attention score matrix is invalid. This insight allows us to guide the $QK$, $Softmax$, and $SV$ operations to skip computations in these invalid regions.

In the $QK$ operation, this principle enables us to directly omit the computation of corresponding results, effectively skipping the red tiling blocks in the result matrix. For the $Softmax$ operation, since it is calculated row by row, we can precisely control which attention scores are included in the computation at the token granularity. This token-level control enables us to support masks of arbitrary shapes while maintaining a mask-free design that eliminates the overhead of users generating and passing in masks externally. For the $SV$ operation, the skipped computations occur during the reduced sum along the $K$ dimension, where certain tiling blocks can be excluded based on the sparsity pattern. By applying these optimizations across the $QK$, $Softmax$, and $SV$ stages, we effectively exploit mask sparsity, ensuring highly efficient and flexible attention mechanisms.

%% file: 5-matmul.tex
\section{SmoothGEMM}
\label{sec_memory}

\begin{figure}[t]
    \centering
    \includegraphics[width=0.95\linewidth]{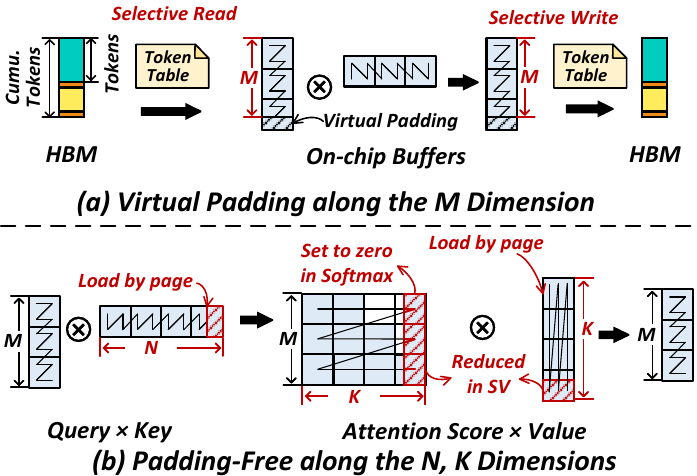}
    \caption{Handling Arbitrary Shapes of GEMM Operations.} 
    \label{fig_padding}
\end{figure}

As discussed earlier, designing a matrix multiplication operation that supports arbitrary shapes while maintaining high performance across all possible shapes is a significant challenge. To address this, we adopt a memory-compute co-design strategy. Instead of optimizing matrix multiplication for every possible shape, we focus on maximizing performance for fixed shapes. To handle arbitrary shapes effectively, we introduce virtual padding at the on-chip memory level, along with selective read and write mechanisms. This approach allows matrix computations to accommodate a wide range of shapes seamlessly while still benefiting from the performance advantages of fixed-shape optimizations.


\subsection{Virtual Padding on the M Dimension}

As illustrated in Fig. \ref{fig_padding}(a), the dimension $M$ in GEMM is intrinsically related to the number of tokens. In the case of Linear GEMM, the $M$ dimension corresponds to the cumulative token count across P/D/V stages, while in Attention GEMM, it is determined by the token count in each stage. In cube-based or tensor-based AI accelerators, matrix computations are typically constrained by a minimum tiling size (e.g., $16 \times 16$ for cube cores). A common practice is to pad the $M$ dimension to align with a multiple of the tiling size. However, this dynamic padding introduces non-trivial memory overhead and degrades performance. 

To mitigate these issues, we replace the physical padding in global memory with virtual padding on the chip, combined with the selective read and write mechanisms shown in Fig. \ref{fig_padding}(a). This approach allows for efficient handling of matrices with arbitrary shapes. Specifically, on-chip buffer allocations are made in tiling-size units to fully exploit the hardware's computational potential. During data transfer from global memory to the on-chip buffer, selective read operations copy only the actual, non-padding data. Similarly, selective write operations ensure that only the non-padding outputs are written back to global memory.

Because the virtual padding regions do not interfere with the computational results of the non-padding regions, this approach guarantees the correctness of the matrix computation results. Moreover, by limiting computations to fixed-shape matrix multiplications, we can apply highly customized optimizations for these fixed shapes, achieving both high efficiency and flexibility for dynamic workloads.

\subsection{Optimizations for N and K Dimensions}

As illustrated in Fig. \ref{fig_moti_gemm}(a), the dimensions of $N$ and $K$ in linear operations are determined by the model structure. These dimensions are typically multiples of hardware tile size, such as $16$, eliminating the need for additional padding.

For Attention GEMM ops, dimensions $N$ and $K$ are tied to the sequence length, which can take arbitrary values. In theory, padding would be required for these dimensions. However, this is naturally handled by the K/V cache’s block page structure, which stores and reads data in blocks aligned to multiples of tile size. As shown in Fig. \ref{fig_padding}(b), by reading entire blocks, the read length inherently conforms to the required hardware tile size. Furthermore, this padding does not affect the final Attention computation because we explicitly set the values in the padded regions to zero during $Softmax$ calculation. This ensures that the padded values are effectively excluded from the Attention score. Additionally, since the padded data is reduced during the Attention $score$ $\times$ $value$ operation, it does not influence the shape of the final output.

Therefore, matrix multiplication for arbitrary shapes can be efficiently supported without introducing additional padding overhead for the $N$ and $K$ dimensions.

\subsection{Handling Dynamic Shapes}
Given our focus on fixed shapes, such as multiples of the tiling size, and operating under budget constraints, we narrow the range of token sizes we handle. While the token count can vary significantly across different workloads, we apply a smoothing technique that reduces the shapes involved to fixed, well-defined sizes. This enables us to perform offline customization and optimization for these specific shapes, particularly for optimizations that are difficult to implement statically, such as swizzling\cite{alexbabey8_optimizing_2020}.

Swizzling is a technique that optimizes matrix multiplication performance by altering the task allocation order across AI cores, enhancing L2 cache hit rates. However, determining the optimal allocation strategy for swizzling is complex and cannot be easily derived through theoretical formulas. To address this challenge, we conduct offline profiling to explore possible access patterns and identify the most efficient inter-core distribution strategy.

These optimal configurations are stored for future use. During online execution, we translate the dynamic workload into the nearest supported fixed shape. Using this fixed shape, we retrieve the pre-optimized configuration from the offline profiling results. This approach ensures that the dynamic workload is handled with the best configuration, enabling optimal performance during execution. By leveraging static optimizations, we can fully exploit the hardware's potential, even in the face of dynamic workloads.
 



\begin{figure}[t]
    \centering
    \includegraphics[width=0.99\linewidth]{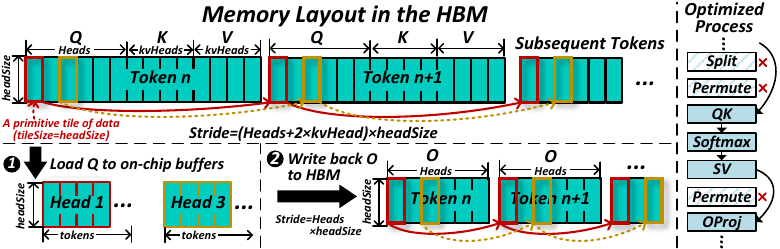}
    \caption{Elimination of Memory Operations.}
    \label{fig-opt-pipe}
\end{figure}

\subsection{Removing Memory Overheads}

For the $QKV$ input, its shape is $[tokens, heads + 2 \times kvHeads, headSize]$, where the outermost dimension corresponds to the tokens, and each token contains its $QKV$ fusion. The attention operation is performed in parallel along the \textit{Head} dimension, which requires the \textit{Head} dimension to be positioned as the outermost dimension. Moreover, we need to read the $Q$, $K$, and $V$ separately rather than the fused $QKV$ tensor. Typically, $Split$ and $Permute$ operations are introduced to reorient the tensor for efficient computation. After the attention computation, another $Permute$ operation is applied to transform the output $O$ back to the shape $[tokens, heads, headSize]$.

To reduce memory overhead, we fuse $Split$ and $Permute$ directly into GEMM computations shown in Fig. \ref{fig-opt-pipe}. For the $QK$ operation, tile-based and stride-based reads are employed to access the required data directly from the $QKV$ tensor, eliminating the need for separate $Split$ and $Permute$ steps. Similarly, for the $SV$ computation, tile-based and stride-based writes are used to store the results in a format that is directly aligned with the $OProj$. This integration removes explicit $Permute$ operations, allowing the $SV$ output to seamlessly flow into subsequent matrix multiplication operations.

%% file: 6-evaluation.tex
\section{Evaluation}
\subsection{XY-Serve Implementation}

We built an Ascend-native inference system based on vLLM\cite{kwon2023efficient}, leveraging Ascend intrinsic to implement core modules such as SmoothGEMM, Meta-Attention, and other essential operators like normalization, activation, and embedding. These operators were exposed to the Python API via pybind11\cite{github_pybind} and seamlessly replaced the corresponding GPU kernels in vLLM, enabling vLLM to support the Ascend NPU.

To reduce the overhead of frequent Python calls, we offloaded the entire model-forward process to C++, integrating the optimized operators into a single C++ function. This model-forward function is then exposed to vLLM for invocation, ensuring a streamlined and efficient execution path.

Additionally, we replaced the native vLLM scheduler with a token-wise scheduling strategy, integrating workload decomposition and computation task reordering to better support dynamic workloads. We redesigned the speculative decoding framework in vLLM, enabling token tree construction and metadata generation for Meta-Attention. These enhancements allow us to implement tree-based speculative decoding algorithms, such as Lookahead Decoding\cite{zhao2024lookahead}, further improving inference performance.

\begin{figure}[t]
    \centering
    \includegraphics[width=0.95\linewidth]{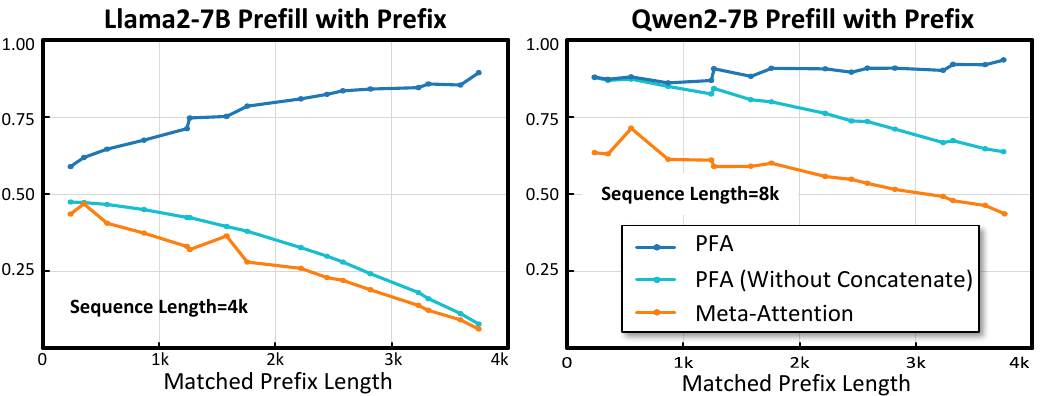}
    \caption{Performance of Prefill Attention with Prefix.}
    \label{fig:evalu_prefix}
\end{figure}

\subsection{Performance of Meta-Attention}

In this section, we evaluate the performance of the attention kernel under dynamic workloads typically encountered in real-world systems. The comparison targets are PromptFlashAttention (PFA) \cite{kernel_PFA} and IncreFlashAttention (IFA) \cite{kernel_IFA} from torch-npu 2.1\cite{torch_npu_ascend}.

\subsubsection{Prefill Attention with Arbitrary Prefix}

In practical systems, the length of the matched system prefix can vary arbitrarily. Therefore, it is crucial to assess performance under arbitrary-length prefix reuse. To simulate this behavior, we adjust the number of reused tokens for an input prompt, token by token, and evaluate performance under different lengths of system prefix matched. Fig. \ref{fig:evalu_prefix} shows the performance under different system prefix reuse (ranging from $0$ to $4k$) for $4k$ and $8k$ prompt inputs. The results show that as the system prefix increases, the processing time of our meta-attention kernel decreases. However, the PFA kernel does not benefit from prefix reuse, primarily because its prefill kernel does not support PagedAttention and only accepts continuous $Q$, $K$, and $V$. When we concatenate the prefix hits from the K/V cache with the new $K$ and $V$, the concatenation time increases as the reuse length grows, counteracting the benefit of prefix reuse. Even when comparing the computation time of the PFA kernel (excluding the K/V concatenation process), our kernel performs an average of $22.4\%$ better.

\subsubsection{Chunked Prefill with Long Sequences}
\begin{figure}[t]
    \centering
    \includegraphics[width=0.95\linewidth]{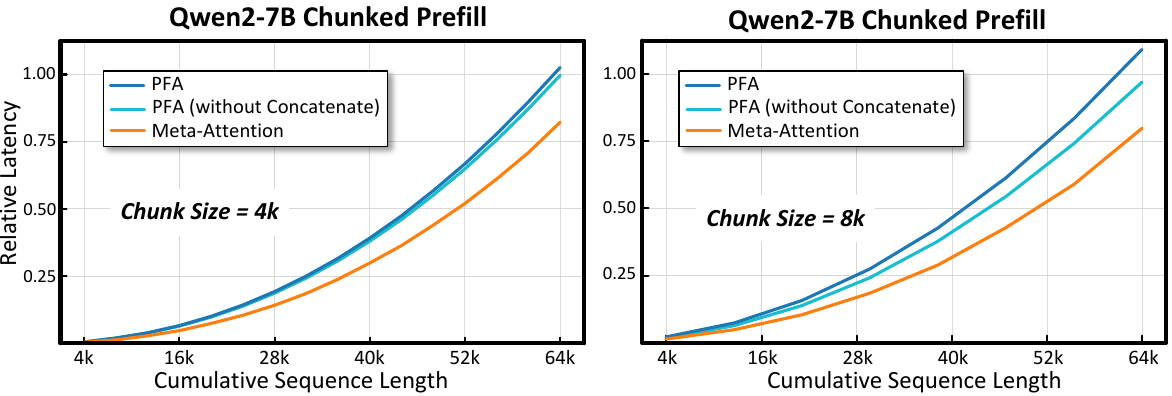}
    \caption{Long Sequence Attention with Chunked Prefill.}
    \label{fig:evalu_chunk}
\end{figure}

For processing long sequences, chunking the sequence into smaller segments is a widely used approach. On the one hand, chunking allows for sequence parallelization by combining it with pipeline parallelism \cite{qin2024mooncake,agrawal2024mnemosyne}. On the other hand, it reduces the impact of prefill on decoding interruptions \cite{agrawal2024taming}. The Chunked Prefill method splits long sequences into multiple chunks, processing them sequentially. After processing each chunk, the corresponding keys and values are stored in the K/V cache for reuse by subsequent chunks. Fig. \ref{fig:evalu_chunk} shows the performance of our Chunked Prefill method for long sequences (with chunk sizes set to $4k$ and $8k$ and a sequence length of up to $64k$). The results demonstrate that our performance surpasses PFA across all sequence lengths. Even when comparing pure computation time with PFA, our kernel shows an improvement up to $22.2\%$.

\begin{figure}[t]
    \centering
    \includegraphics[width=0.95\linewidth]{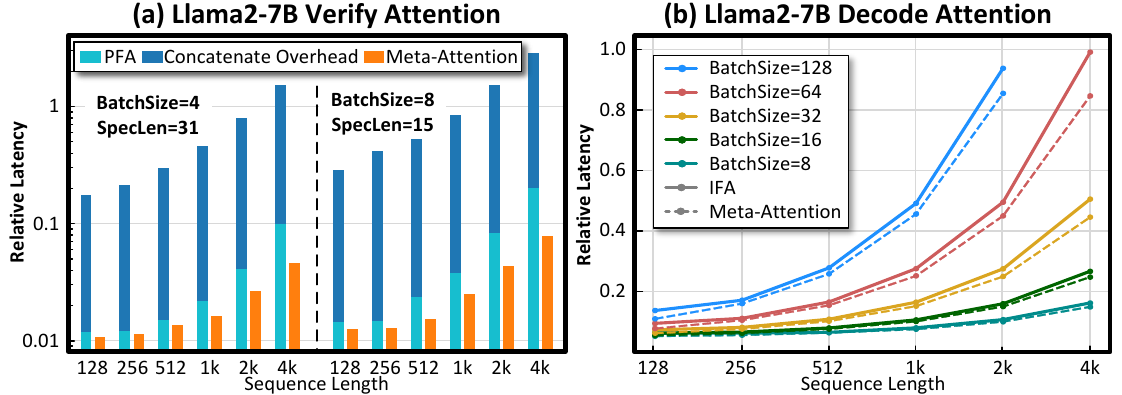}
    \caption{Performance of LLM Verify and Decode Attention.}
    \label{fig:evalu_decode}
\end{figure}

\subsubsection{Speculative Decoding}
Next, we evaluate the performance of the verify kernel under different context lengths. We compare the performance with a $batchSize=4$, $specLen=31$ and a $batchSize=8$, $specLen=15$. Fig. \ref{fig:evalu_decode}(a) shows that, across different context lengths, our kernel consistently outperforms PFA. Furthermore, as the context length increases, the performance improvement becomes increasingly time-consuming as the historical length grows. Even when excluding this concatenation operation from PFA and comparing only the pure computation time, our kernel still demonstrates an average improvement of $28.6\%$.

\subsubsection{Decode Performance}
In the decode phase, both the context length and batch size can vary arbitrarily. To flexibly support decoding with arbitrary context lengths, we enable PagedAttention optimization. To evaluate decoding performance under such conditions, we measure performance across different context lengths and batch sizes. Fig. \ref{fig:evalu_decode} presents the performance of Llama2-7B under varying sequence lengths and batch sizes. The results show that, compared to the IFA kernel, our kernel achieves performance improvements across all combinations of batch size and sequence length, with average $12.9\%$ improvements.

\begin{figure}[t]
    \centering
    \includegraphics[width=0.95\linewidth]{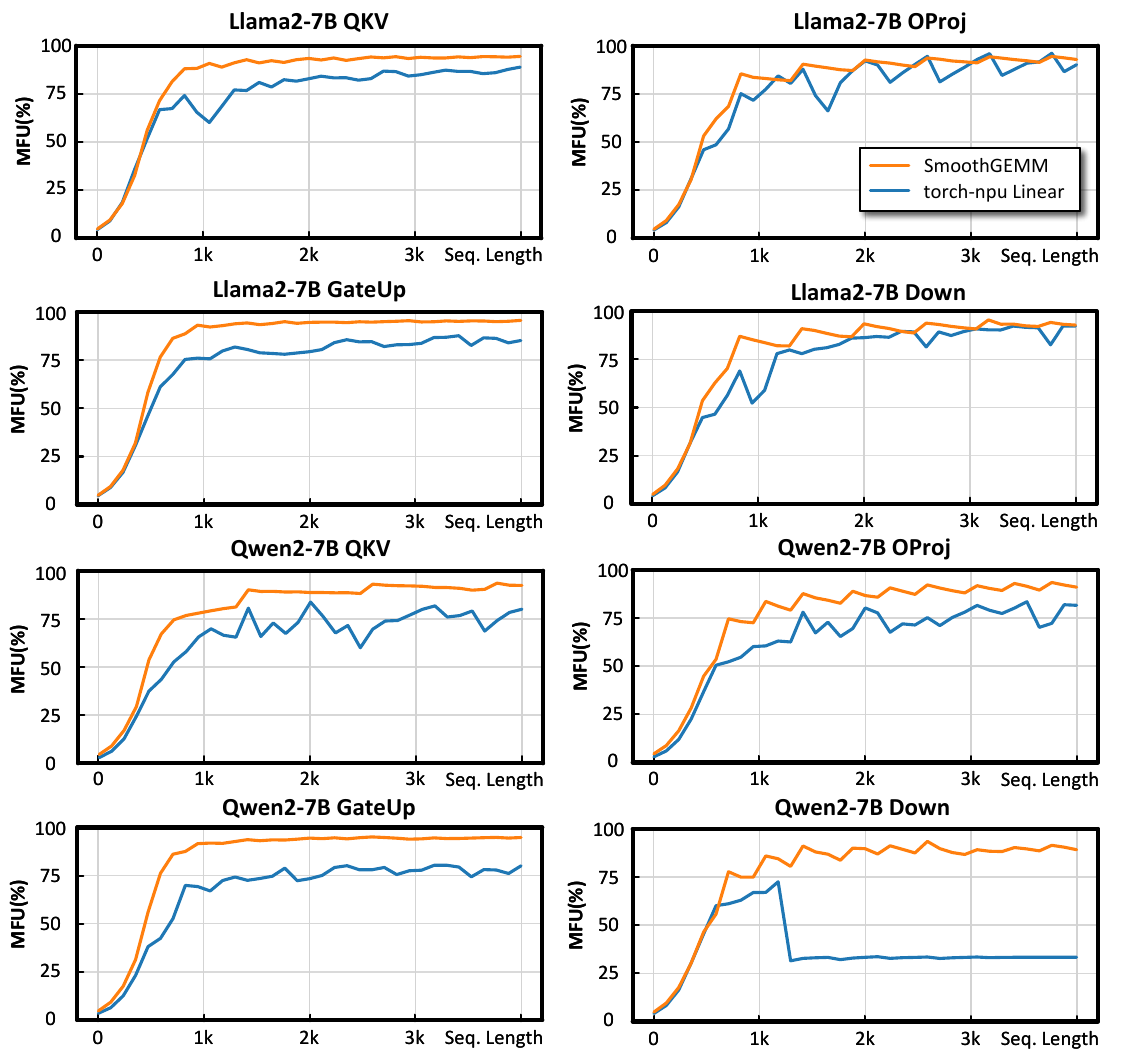}
    \caption{Linear GEMM Performance.}
    \label{fig:evalu_matmul}
\end{figure}

\subsection{Performance of SmoothGEMM}
In practical systems, the input length from users can vary arbitrarily, ranging from $0$ to the maximum length supported by the model. For long sequences, to minimize the impact of prefill on decode and to optimize sequence parallelization, we typically adopt the Chunked Prefill strategy, which imposes a constraint on the maximum chunk length, such as $4k$. In real-world scenarios, lengths smaller than the chunk size may also be encountered. To evaluate performance across different conditions, we assess the LLMs with input lengths ranging from $1$ to the chunk size ($4096$).

We compare the performance of linear operators ($QKV$, $OProj$, \emph{GateUp}, and $Down$) using shapes derived from Llama2-7B and Qwen2-7B with TP=$1$. As displayed in Fig. \ref{fig:evalu_matmul}, the results indicate that SmoothGEMM outperforms torch-npu linear by an average of $14.6\%$, demonstrating superior performance across nearly all tested shapes. Moreover, the performance remains stable, with ideal MFU typically achieved for sequence sizes above $1k$.

\subsection{End-to-End Evaluation}

\subsubsection{vLLM Nightly Benchmarks}

\begin{figure}[t]
    \centering
    \includegraphics[width=0.99\linewidth]{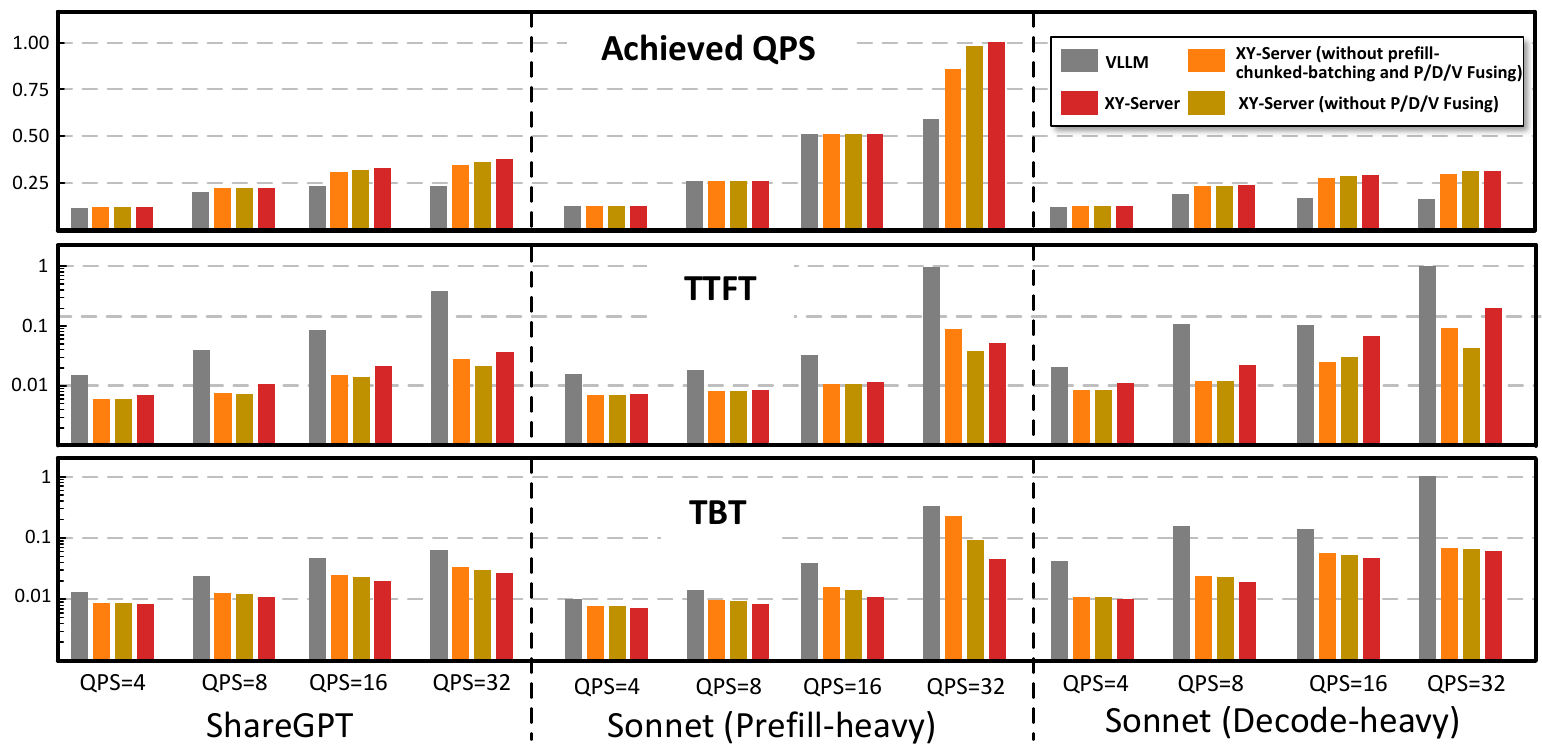}
    \caption{End-to-End Evaluation on Nightly Benchmarks.}
    \label{fig:evalu_endtoend}
\end{figure}

For end-to-end benchmarking, we use the nightly-benchmarks from the vLLM community \cite{nightly-benchmark}, with Qwen2-7B as the model. The baseline comparison is against a community version of vLLM that supports Ascend NPU (Ascend-vLLM) \cite{vLLM-for-NPU}, primarily utilizing GEMM provided by torch-npu and Fused-Attention operators for computation. The test data is divided into three scenarios: ShareGPT, Prefill-heavy(462 input tokens and 16 output tokens), and Decode-heavy(462 input tokens and 256 output tokens).

As shown in Fig. \ref{fig:evalu_endtoend}, we measure performance under fixed request rates per second (QPS) of $4$, $8$, $16$, and $32$ for each test dataset. The following metrics are collected: average Time-to-First-Token (TTFT), average Time-between-Tokens (TBT), and achieved QPS. Even without enabling advanced features such as prefill-chunked-batching and P/D/V fusing, XY-Serve demonstrates a clear performance advantage over the baseline. Specifically, XY-Serve achieves an achieved QPS improvement of up to $79\%$ across various workload types. Additionally, it delivers $64\%$ lower average TTFT and $57\%$ lower average TBT latency. This improvement is primarily attributed to the efficient optimization of operator implementations.

With the dynamic scheduling optimizations of prefill-chunked-batching and PDV fusing enabled, XY-Serve further gains an achieved QPS improvement up to $89\%$ and reduces average TBT latency by $69\%$ across all scenarios. This outcome underscores XY-Serve's strong support for dynamic workloads, effectively benefiting from these enhancements.

When prefill-chunked-batching is enabled, the length of tokens processed in each prefill is effectively maintained at the budgeted length, improving MFU and reducing TTFT under high-pressure conditions. However, enabling P/D/V fusing results in a slight deterioration of TTFT latency in the Decode-heavy scenario under high throughput pressure. In this case, the number of decodes fused with the prefill increases, which slightly impacts the TTFT.




\subsubsection{Ascend NPUs VS. GPUs}
We compared the end-to-end inference MFU and MBU between XY-Serve running on the 910B and the official vLLM-v0.6.4.post1 on the Nvidia A800. The measurements were taken during the prefill and decode stages of the entire forward pass for the Qwen2-7B and Llama2-7B models at TP=$1$, across various sequence lengths. As shown in Fig. \ref{fig:evalu_gpu}, XY-Serve achieves MFU and MBU similar to the A800. Notably, in terms of MBU, XY-Serve demonstrates a clear advantage over GPUs, showing an improvement up to $17\%$.

%% file: 7-relatedwork.tex
\section{Related Work}

\textbf{Attention Optimization: } FlashAttention1\cite{dao2022flashattention} and FlashAttention2\cite{dao2023flashattention} optimize the prefill phase by tiling computations to avoid HBM access, improving performance. FlashAttention3\cite{shah2024flashattention} further enhances performance through parallelism between softmax and matrix operations. FastAttention\cite{lin2024fastattention} extends FlashAttention2 from GPUs to Ascend NPUs, while FlashDecoding\cite{_flashdecoding_} improves decoding efficiency for small batches by splitting along the sequence dimension. Recent works\cite{_accelerating_, ye-etal-2024-chunkattention} further optimize decoding performance by transforming GEMV operations into GEMM operations when sharing prefixes. While these techniques primarily target either the prefill or decode phases, POD-Attention\cite{kamath2024pod} simultaneously optimizes both, maximizing computational power and bandwidth. In contrast, our work tackles real-world deployment scenarios with dynamic prefix reuse and speculative algorithms, leading to mixed P/D/V stages. We decompose dynamic workloads into hardware-friendly meta-primitives, simplifying attention module design.

\textbf{Linear Optimization: }
Existing techniques like swizzling\cite{alexbabey8_optimizing_2020}, split-k\cite{_accelerating_a}, and ping-pong\cite{_deep_} are commonly used in linear optimization. Our approach shows that supporting specific matrix shapes is sufficient for dynamic LLM workloads. By optimizing these shapes offline using the above techniques, we store the configurations and apply them during online execution to achieve optimal performance.

\textbf{Serving Systems: }
Several works aim to address the dynamic nature of inference systems. Orca\cite{yu2022orca} mitigates output length variability through iteration-level scheduling, while PageAttention\cite{kwon2023efficient} optimizes memory allocation for KV caches, reducing waste from fixed-length allocations. Our approach builds on these, tackling additional complexities in real-world inference systems, such as hybrid P/D/V stages and dynamic shapes in each stage.

The interruption of prefill on decode can increase TBT. Two strategies have been proposed to address this: SplitFuse\cite{holmes2024deepspeed, agrawal2023sarathi, agrawal2024taming}, which divides the Prefill phase into smaller chunks and fuses chunks with decode and disaggregated LLMs\cite{hu2024inference, qin2024mooncake, zhong2024distserve, jin2024p}, which separate Prefill and Decode across different machines. XY-Serve supports both two deployments. It also enables seamless transitions between Prefill and Decode roles in disaggregated setups.

\begin{figure}[t]
    \centering
    \includegraphics[width=0.95\linewidth]{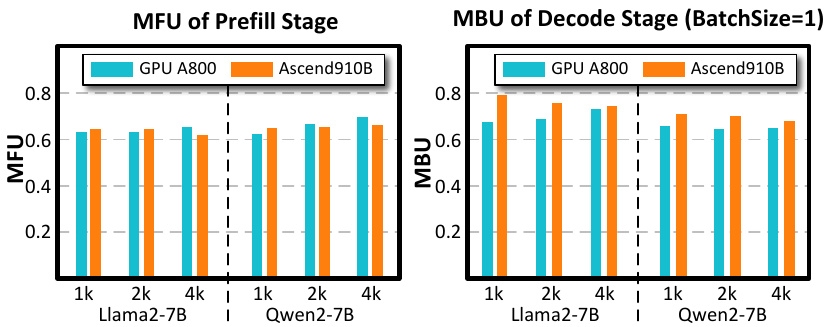}
    \caption{Comparison between Ascend NPUs and GPUs.}
    \label{fig:evalu_gpu}
\end{figure}

%% file: 8-conclusion.tex
\section{Conclusion}
We introduced XY-Serve, an end-to-end production serving system designed to tackle challenges of dynamic LLM workloads. By integrating workload decomposition, computation task reordering, and meta kernels (Meta-Attention and SmoothGEMM), XY-Serve optimizes throughput, latency, and computational efficiency in real-world production environments. Experimental results demonstrate superior performance on Ascend NPUs, delivering MFU and MBU comparable to GPUs. With its flexibility to handle diverse dynamic workloads, XY-Serve sets a new benchmark for efficiency and adaptability in production-grade LLM inference systems.